\documentclass[11pt]{article}
\usepackage{graphicx} 

\usepackage[utf8]{inputenc}

\usepackage{fullpage}
\usepackage[margin = 2.5cm]{geometry}

\usepackage{amsmath, amsthm, amssymb}
\usepackage{thmtools}
\usepackage{thm-restate}
\usepackage{mathtools}  
\usepackage{xfrac} 

\usepackage{placeins} 
\usepackage{times}

\usepackage{url}
\usepackage{array}

\usepackage[ruled,vlined]{algorithm2e}
\newenvironment{algorithm2e}[1][]{%
    \begin{algorithm}[#1]%
}{%
    \end{algorithm}
}

\usepackage{graphicx}
\usepackage{color}

\usepackage[round]{natbib}
\usepackage[hyperindex,breaklinks]{hyperref}
\usepackage{url}

\usepackage{tcolorbox}

\usepackage{tikz}

\usepackage{nicefrac}

\newtheorem{theorem}{Theorem}[section]

\newtheorem{corollary}[theorem]{Corollary}

\newtheorem{lemma}[theorem]{Lemma}
\newtheorem{definition}[theorem]{Definition}

\newtheorem{assumption}[theorem]{Assumption}
\newtheorem*{remark*}{Remark}

\usepackage{cleveref}
\crefname{theorem}{theorem}{theorems}
\Crefname{theorem}{Theorem}{Theorems}

\crefname{lemma}{lemma}{lemmas}
\Crefname{lemma}{Lemma}{Lemmas}

\crefname{claim}{claim}{claims}
\Crefname{claim}{Claim}{Claims}

\crefname{corollary}{corollary}{corollaries}
\Crefname{corollary}{Corollary}{Corollaries}

\crefname{proposition}{proposition}{propositions}
\Crefname{proposition}{Proposition}{Propositions}

\crefname{definition}{definition}{definitions}  
\Crefname{definition}{Definition}{Definitions}

\crefname{observation}{observation}{observations}
\Crefname{observation}{Observation}{Observations}

\crefname{question}{question}{questions}
\Crefname{question}{Question}{Questions}

\crefname{assumption}{assumption}{assumptions}
\Crefname{assumption}{Assumption}{Assumptions}

\crefname{algorithm}{algorithm}{algorithms}
\Crefname{algorithm}{Algorithm}{Algorithms}

\crefname{AlgoLine}{line}{lines}  
\Crefname{AlgoLine}{Line}{Lines}

\DeclareMathOperator{\one}{\mathbf{1}}

\newcommand{\bbR}{\mathbb{R}}

\newcommand{\calH}{\mathcal{H}}

\newcommand{\calO}{\mathcal{O}}

\newcommand{\calS}{\mathcal{S}}

\newcommand{\N}{\mathbb{N}}
\newcommand{\R}{\mathbb{R}}

\DeclareMathOperator{\Tr}{Tr}

\newcommand{\inner}[2]{\langle #1, #2\rangle}

\title{Verifying Classification with Limited Disclosure}
\author{Siddharth Bhandari, Liren Shan}
\date{}

\begin{document}

\maketitle

\begin{abstract}%
  We consider the multi-party classification problem introduced by Dong, Hartline, and Vijayaraghavan (2022) motivated by electronic discovery. In this problem, our goal is to design a protocol that guarantees the requesting party receives nearly all responsive documents while minimizing the disclosure of nonresponsive documents. We develop verification protocols that certify the correctness of a classifier by disclosing a few nonresponsive documents. 
  
  We introduce a combinatorial notion called the \emph{Leave-One-Out} dimension of a family of classifiers and show that the number of nonresponsive documents disclosed by our protocol is at most this dimension in the realizable setting, where a perfect classifier exists in this family. 
  For linear classifiers with a margin, we characterize the trade-off between the margin and the number of nonresponsive documents that must be disclosed for verification. Specifically, we establish a trichotomy in this requirement:
  for $d$ dimensional instances, when the margin exceeds $1/3$, verification can be achieved by revealing only $O(1)$ nonresponsive documents; when the margin is exactly $1/3$, in the worst case, at least $\Omega(d)$ nonresponsive documents must be disclosed; when the margin is smaller than $1/3$, verification requires $\Omega(e^d)$ nonresponsive documents.
  We believe this result is of {independent interest} with applications to {coding theory} and {combinatorial geometry}.
  We further extend our protocols to the nonrealizable setting defining an analogous combinatorial quantity robust Leave-One-Out dimension, and to scenarios where the protocol is tolerant to misclassification errors by Alice.
\end{abstract}

\section{Introduction}

Machine learning is increasingly being used to automate decision-making, enabling faster and more efficient information processing. 
In particular, classification models play a crucial role in efficiently identifying relevant information while minimizing manual review efforts.
However, in sensitive domains such as legal discovery, financial auditing, and medical diagnostics, a critical challenge is ensuring that classification models maintain high accuracy while preserving privacy.

To address this challenge, we focus on the classification problem in the context of legal discovery.
In the \emph{legal discovery process}, the plaintiff (Bob) issues a request for production to the defendant (Alice) seeking relevant evidence from the documents that Alice possesses. 
Alice is then accountable for identifying and providing the responsive (relevant) documents to Bob. 
To efficiently process massive amounts of electronic documents, technology-assisted review (TAR) tools are widely used in legal discovery to retrieve responsive documents with significantly less human review~\citep{grossman2010technology}. 
Despite its advantages, the adoption of TAR involves a potential concern that it may reduce accountability and transparency due to model bias, insufficient training data, or adversarial manipulation so as to hide unfavorable responsive documents.
To ensure the accuracy of Alice's classification, Bob may need to review a subset of documents, including nonresponsive ones. However, this process risks exposing private information from nonresponsive documents.
Thus, this raises a fundamental question: 
\emph{How can we verify classification correctness while minimizing the disclosure of nonresponsive documents?}

Several straightforward approaches could be considered \citep*{dong2022classification}: placing the burden of classification on Alice and holding her accountable; requiring Alice to disclose all documents to Bob; having Alice reveal all documents along with a classification mechanism (supplied by Bob) to a trusted third party, such as Trent (a court system or cloud provider); or asking the court to adjudicate the relevance of all documents.  
However, due to concerns regarding accountability, privacy, and efficiency, none of these approaches adequately address the fundamental question.  
Placing the burden on Alice makes classification accuracy dependent on her accountability and ethical obligations. This misaligns incentives, as her legal team is expected to assist an adversary, conflicting with their role~\citep{gelbach2015law}. Requiring Alice to disclose all documents compromises privacy, as nonresponsive documents may contain sensitive information. While confidentiality agreements exist, they are unreliable between untrusted parties. Expecting Bob to provide a meaningful classifier is impractical. He may struggle to encode a labeling strategy or require access to Alice’s documents. Worse, he could supply a corrupt classifier to extract sensitive information. Court adjudication exposes nonresponsive documents, compromising privacy, and incurs significant legal costs and delays.
Given these limitations, a more nuanced solution is necessary to balance accountability and privacy, while being efficient.

In this regard, \cite{dong2022classification} recently introduced the multi-party classification problem for electronic discovery (e-discovery) in legal proceedings. 
They developed a multi-party protocol, using a trusted third party, that guarantees Bob receives all the responsive documents by verifying a minimal subset of nonresponsive documents. 
Thus, this protocol addresses the accountability issue while minimizing Alice's privacy loss. It has the further advantage of being computationally efficient and revealing all the responsive documents turns out to be Alice's best strategy.
However, their protocol has two limitations. 
First, it is designed specifically for linear classification (including kernel methods) in a realizable setting, where responsive and nonresponsive documents can be perfectly separated by a linear decision boundary. 
In contrast, real-world document embeddings or classifications produced by language models (such as LLMs) often result in complex decision boundaries that may not be linear or even realizable by a specific family of classifiers.
Recent work of \citet*{dong2024error} considers the nonrealizable setting, but only for linear classifiers with one-dimensional embedding.
Second, although the protocol by~\cite{dong2022classification} identifies a minimal set of nonresponsive documents necessary for verification, in the worst case, all documents are disclosed to Bob, leading to a complete exposure of private information in nonresponsive documents during verification.

In this work, we address the above issues. 
Firstly, we extend the multi-party classification protocol to a general family of classifiers with arbitrarily complex decision boundaries.
A crucial subroutine in \citet{dong2022classification} was a protocol for computing critical points, a concept we clarify later in the paper. This is also essential in our setting.  
However, there is a \emph{key difference in our critical points protocol}, which enables us to extend to arbitrary family of classifiers.
Further, our protocol is efficient provided a membership oracle for the classifier family.
Secondly, we show that for a given family of classifiers, verifying the correctness of a classification requires disclosing only a limited number of nonresponsive documents. This number is characterized by a combinatorial measure, which we call the \emph{Leave-One-Out dimension}.
Hence, for certain families of classifiers we can significantly improve privacy guarantees while guaranteeing that Bob learns all the responsive documents. 
We establish nontrivial bounds on the Leave-One-Out dimension for linear classifiers with sufficient margin, using an independently interesting geometric lemma involving a skew-orthogonal family of vectors.
Further, in the worst case (in a sense made precise later) the Leave-One-Out dimension also serves as a lower bound on the number of nonresponsive documents needed to be revealed so as to guarantee that Bob learns all the responsive documents.
Thirdly, we also analyze the non-realizable setting and show that a similar protocol is able to achieve accountability while minimizing Alice's privacy loss.
Here a robust version of the Leave-One-Out dimension plays the analogous role.
Finally, we demonstrate a protocol which works in a general non-realizable setting and only leaks a few nonresponsive documents per error in classification committed by Alice. This helps avoid undesirable feature in all previous protocols of revealing all the documents once a certain number of classification errors were detected.


\subsection{Model}

We now outline the framework for verifying classification in a multi-party setting, as introduced by~\cite{dong2022classification}.
In this setting, the defendant (Alice) possesses a set of documents embedded in a $d$-dimensional space, $X \subset \mathbb{R}^d$. The plaintiff (Bob) issues a request for production, seeking documents responsive to his request. Let $f: X \to \{-1,+1\}$ denote the true labeling function, where a document $x \in X$ is considered responsive if $f(x) = 1$ and nonresponsive otherwise.
The pair $(X, f)$, consisting of the document set and its true labels, is referred to as an instance.

A multi-party protocol for verifying classification involves three parties: Alice, Bob, and a trusted third party, Trent. Trent may be represented by the court system or a cloud computing service provider.\footnote{For further details about this assumption please see \cite*{dong2022classification}[Introduction].}  
Alice and Bob have the ability to label documents but only Alice holds $X$.
The protocol unfolds through multiple rounds of interaction among these parties.
While Trent has access to all documents and can perform computations on them, they cannot assign labels.
Instead, Trent facilitates the process by requiring Alice and Bob to review and label specific documents.
When necessary, the court may be called upon to adjudicate the true label of a document. However, this is considered an expensive operation due to the time and cost associated with legal proceedings and should therefore be used sparingly.
Furthermore, if the true label is nonresponsive this would also count as a nonresponsive disclosure, which hurts Alice's privacy.
At the conclusion of the protocol, a subset of documents, $B \subseteq X$, is disclosed to Bob.
Throughout the execution of the protocol, Alice and Bob may deviate maliciously in their best interest.

We evaluate the performance of a protocol using the following metrics as in \citet*{dong2024error}. 
\begin{definition}[Recall and Nonresponsive Disclosure]
Let $(X,f)$ be an instance, and let $n^+ = |\{x \in X : f(x) = 1\}|$ denote the number of responsive documents.

The \emph{Recall} of a protocol is the worst-case fraction of responsive documents retrieved and disclosed to Bob (over Alice's responses), assuming Bob acts faithfully:
\[
\mathrm{Recall} = \frac{|\{x \in B : f(x) = 1\}|}{n^+}.
\]

The \emph{Nonresponsive Disclosure} quantifies Alice’s privacy loss in the worst case (over Bob’s responses). It is defined as the number of nonresponsive documents revealed to Bob, including those adjudicated by the court, assuming Alice reports her labels correctly:
\[
\mathrm{Nonresponsive\ Disclosure} = |\{x \in B : f(x) = -1\}|.
\]
\end{definition}

Then, our goal is to design a protocol that achieves a Recall as close to $1$ as possible while minimizing the Nonresponsive Disclosure.

\subsection{Our Results}

In this work, we provide multi-party protocols that verify the correctness of a classification while minimizing the disclosure of nonresponsive documents in various settings.

A crucial quantity in our analysis is the following combinatorial concept, called the \emph{Leave-One-Out} dimension.
The Leave-One-Out dimension is closely related to the star number\footnote{See Definition 2 in Section 4 of \cite{hanneke2015minimax}.} introduced by~\cite{hanneke2015minimax}.
A set system consists of a set $X$ and a collection $\calS$ of subsets of $X$. 
We first define the following Leave-One-Out dimension of a set system. 

\begin{definition}[Leave-One-Out dimension]
\label{defn:leave_one_out}
     Given a set system $(X,\calS)$, the \emph{Leave-One-Out} dimension is the cardinality of the largest set $C\subseteq X$ such that for each element $c \in C$, there exists $S \in \calS$ with $S \cap C = \{c\}$.
\end{definition}

Consider a class of binary classifiers on $X$, denoted by $\calH = \{f:X \to \{-1,+1\}\}$. We define the associated set family as $\calS = \{f^+ : f \in \calH\}$ where $f^+ = \{x: f(x) = 1\}$ is the set of elements classified as positive by $f$. The \emph{Leave-One-Out} dimension of this classifier class $\calH$ is defined as the Leave-One-Out dimension of the set family $\calS$. This means there exists a set $C$ with size same as the Leave-One-Out dimension of $\calH$ such that for each element $c \in C$, a classifier $f \in \calH$ classifies this element as positive and all other elements  of $C$ as negative.   

Given a class $\calH$ of binary classifiers on $X$, an instance $(X,f)$ is realizable by this class if there exists a classifier $h \in \calH$ that perfectly classifies this instance, i.e. $h(x) = f(x)$ for any $x\in X$. Otherwise, this instance is called nonrealizable.

\textbf{Realizable Setting.} 
We first consider the realizable setting, where the true labels can be perfectly separated by a classifier from a general hypothesis class with an arbitrary decision boundary. We develop a multi-party classification protocol that verifies classification  correctness with perfect recall, ie, $1$, and with a nonresponsive disclosure equal to the Leave-One-Out dimension of the hypothesis class. 

Further Alice’s best strategy is to report labels truthfully. Moreover, the protocol remains efficient, assuming access to an oracle $\mathcal{O}$ that verifies label membership in the hypothesis class. 
Additionally, we show that any protocol with perfect recall, i.e., $\mathrm{Recall} = 1$, incurs a nonresponsive disclosure of at least one less than the Leave-One-Out dimension.
See \Cref{thm:protocol_realizable} for further details.

\textbf{Linear Classification with Margin.}
We then consider the special case where the instance is realizable by the class of linear classifiers with a margin (see \Cref{defn:margin}).
We establish a fundamental trade-off between margin size and the number of nonresponsive disclosures required for verification. 
Specifically, for any instance in $\mathbb{R}^d$ that is linearly separable with margin $\gamma$, we show a trichotomy in the nonresponsive disclosure/Leave-One-Out dimension in Table~\ref{tab:margin_tradeoff}. See \Cref{thm:Leave-One-Out-linear-realizable} for more details.
Our protocol is efficient for the class of linear classifiers with a margin since there is an efficient oracle $\calO$ that verifies the label membership in this class using the hard support vector machine (SVM) (See Theorem 15.8 in~\citet*{shalev2014understanding}).

\FloatBarrier
\begin{table}[ht]
    \centering
    \renewcommand{\arraystretch}{1.3}
    \begin{tabular}{|m{4.8cm}|m{2cm}|m{2cm}|m{4cm}|}
        \hline
        \textbf{Margin $\gamma$} & \textbf{$\gamma > 1/3$} & \textbf{$\gamma = 1/3$} & \textbf{$\gamma < 1/3$} \\
        \hline
        \textbf{Nonresponsive Disclosures (Leave-One-Out dimension)}  & $\frac{2 + 2\gamma}{3\gamma - 1}$ & $\Omega(d)$ & $\exp(\Omega((1/3-\gamma)^2d))$ \\
        \hline
    \end{tabular}
    \caption{Trade-off between margin size and nonresponsive disclosures (Leave-One-Out dimension) for instances in \(\mathbb{R}^d\) that are linearly separable with margin \(\gamma\).}
    \label{tab:margin_tradeoff}
\end{table}
\FloatBarrier

\begin{remark*}
    The above is a geometric statement about a skew-orthogonal family of vectors (see \Cref{lemma:skew-obtuse_lemma}) which we believe is of independent interest in combinatorial geometry and coding theory.
\end{remark*}

Hence, for instances linearly separable with margin $\gamma > 1/3$, our protocol verifies classification with at most 
$\frac{2 + 2\gamma}{3\gamma - 1}$
nonresponsive disclosures, independent of the total number of documents and dimension $d$. This result highlights the privacy benefits of using classifiers with large margins in the multi-party classification.
In the specific case of $d=1$ the Leave-One-Out dimension turns out to be $2$, since we can always assume $\gamma = 1$.

\textbf{Nonrealizable Setting.}
We then extend our protocols to the nonrealizable setting where the true labels can not be perfectly classified by any classifier in the hypothesis class.
In this setting, we introduce a robust verification protocol (see \Cref{sec:non_realizable}) that achieves a recall loss \emph{at most the error rate of the optimal classifier} in the hypothesis class with a nonresponsive disclosure equal to the \emph{Robust-Leave-One-Out} dimension. This parameter is a robust analogue of the Leave-One-Out dimension. See~\Cref{defn:robust_leave_one_out} for more details. 


For linear classifiers, we prove that even in the nonrealizable setting, the Robust-Leave-One-Out dimension exhibits the same trichotomy as the Leave-One-Out dimension, albeit nuanced by the error rate of the optimal linear classifier. Specifically, if there exists a linear classifier with margin $\gamma > \frac{1}{3}$ and at most $L$ misclassifications, the nonresponsive disclosure is at most 

\[
\frac{(2+2L)(\gamma+1)}{3\gamma -1}.
\]

Thus, even in the nonrealizable setting, the nonresponsive disclosure remains independent of the dimension as long as the margin exceeds $\frac{1}{3}$.

\textbf{Error-Tolerant Protocol.} We also extend our protocols to a protocol that is tolerant of misclassification errors by Alice. 
In the above protocols, detecting a single misclassification made by Alice led to the disclosure of all documents to Bob. 
However, in practice, even the best human reviewer may make unintended mistakes in review tasks.
To address this, we convert our protocols to error-tolerant protocols that ensure the large recall guarantee while maintaining a small nonresponsive disclosure that scales with the number of errors made by Alice. The scaling factor is the Leave-One-Out or the Robust-Leave-One-Out dimension depending on the setting.
See \Cref{thm:error_tolerant_protocol} for more details.

\subsection{Related Work}

\cite{dong2022classification} introduced a multi-party classification problem and designed the critical points protocol that ensures that all responsive documents are disclosed while revealing as few nonresponsive documents as possible. 
Their approach relies on linear classification and the realizable setting where there exists a linear classifier that correctly classifies all documents. 
Later, \citet{dong2024error} extended this problem to the non-realizable setting, where all documents may not be perfectly classified by a linear classifier. 
They developed a protocol that for one-dimensional embedding of documents, ensures a recall loss at most the error rate of the optimal linear classifier, and discloses at most $O(\log n)$ nonresponsive documents. They further proposed a heuristic-based protocol for high-dimensional embedding, demonstrating its effectiveness through empirical evaluations.

Linear classification is widely studied in machine learning literature. 
In the realizable setting, where data is perfectly separable by a linear classifier, passive learning can achieve an error rate $\varepsilon$ using empirical risk minimization (ERM) with a sample complexity of $O(d/\varepsilon)$ on $d$ dimensional space~\citep{vapnik1998statistical}. Compared to passive learning, active learning, where the learner adaptively selects which data points to label, can significantly reduce the label complexity. Disagreement-based methods by~\citet*{balcan2006agnostic, hanneke2007bound} can achieve $\varepsilon$ error rate with $O(\Theta d\log (1/\varepsilon))$ queries where $\Theta$ is the disagreement coefficient.
For one-dimensional instances, to achieve a $1/n$ error rate, the passive learning and the active learning require $O(n)$ and $O(\log n)$ queries respectively, while verification only requires one query. For $d > 1$, in the worst case, both passive learning and active learning require $O(n)$ queries since the disagreement coefficient can be $1/n$.
Learning linear classifiers in the nonrealizable setting is known to be computationally hard in high dimensions~\citep*{kearns1994introduction,kalai2008agnostically,guruswami2009hardness}. 

\citet*{goldwasser2021interactive} studied a distinct two-party classification problem known as PAC verification, which is motivated by the delegation of computation. The goal in PAC verification is for the prover to convince the verifier that a given classifier is approximately correct, using significantly fewer labeled examples than would be required for the verifier to learn the classifier independently.
A key difference between their setting and ours is that in PAC verification, both the prover and verifier have access to the distribution of labeled data. In contrast, in our framework, only Alice has access to the labeled data distribution.

\section{Realizable Setting}

In this section, we consider the instance $(X,f)$ that is realizable by some hypothesis class $\calH$. 
In Section~\ref{sec:realizable_general}, we provide a multi-party verification  protocol for the general hypothesis class that achieves perfect recall and nonresponsive disclosure at most the Leave-One-Out dimension (Definition~\ref{defn:leave_one_out}) of the hypothesis class.
In Section~\ref{sec:realizable_linear_margin}, we characterize the Leave-One-Out dimension for the family of linear classifiers with margin $\gamma$.


\begin{theorem}\label{thm:protocol_realizable}
    Let $\calH$ be a hypothesis class of binary classifiers on a set $X$ with Leave-One-Out dimension $k$.  
    Suppose $f: X \to \{+1,-1\}$ represents the true labels for responsive and nonresponsive documents, and let $(X,f)$ be a realizable instance with respect to $\calH$. Then, \Cref{alg:abstract_critical_points_protocol} defines a multi-party verification protocol with the following properties:
    \begin{enumerate}
        \item \textbf{(Recall)} The recall is $1$.
        \item \textbf{(Nonresponsive Disclosure)} If Alice reports all labels correctly, the number of disclosed nonresponsive documents is at most $k$.
        \item \textbf{(Truthfulness)} Alice's best strategy is to report all labels truthfully.
        \item \textbf{(Efficiency)} Given an oracle $\calO$ for membership testing in $\calH$, the protocol runs in time $O(|X|)$.
    \end{enumerate}
    Furthermore, there exists a subset $X' \subseteq X$ such that $(X', f)$, with $f(x) = -1$ for all $x \in X'$, is realizable with respect to $\calH$ (when restricted to $X'$), and any protocol achieving recall $1$ on $(X', f)$ must incur a nonresponsive disclosure of $k-1$.

\end{theorem}

\subsection{Protocol for General Hypothesis Class}\label{sec:realizable_general}

Consider the  multi-party verification protocol \Cref{alg:abstract_critical_points_protocol} when the input is an instance $(X,f)$, realizable by a general hypothesis class $\calH$. The protocol contains a subroutine for computing critical points as shown in Algorithm~\ref{alg:abstract_critical_points_computation}. 
This protocol makes $O(|X|)$ queries to an oracle $\calO$ that checks whether a set of labeled points is realizable by the class $\calH$.
This protocol works as follows.
First, Alice provides Trent with the entire set of documents $X$ along with her labels. Next, Trent applies Algorithm~\ref{alg:abstract_critical_points_computation} to identify critical points based on Alice’s labeling. 
In this algorithm, Trent iterates over each document labeled by Alice as negative, temporarily flips its label to positive, and checks—using an oracle—whether any classifier in $\calH$ can perfectly classify the resulting dataset. If no such classifier exists, that document is removed from consideration. All remaining negatives after this procedure are deemed critical points. Finally, the protocol sends all documents labeled as positive and all critical points to Bob for verification.

There is a key difference between our protocol and the critical points protocol from~\cite{dong2022classification}. In our protocol, if no classifier in $\calH$ perfectly classifies $T_1$ and $T_2$, we remove a point in Step 5. 
By contrast, the protocol in~\cite{dong2022classification} does not remove any points. 
This distinction allows our protocol to handle a general hypothesis class $\calH$. 
To illustrate, consider a classifier $h \in \calH$ that labels $X_A^+$ and two additional points $x_1,x_2 \in X_A^-$ as positive and labels all remaining points $X \setminus (X_A^+\cup \{x_1,x_2\})$ as negatives. 
If there is no classifier in $\calH$ that labels exactly $X_A^+ \cup \{x_1\}$ as positive or exactly $X_A^+ \cup \{x_2\}$ as positive, then the protocol in~\cite{dong2022classification} will not mark $x_1$ or $x_2$ as critical points. Consequently, it can not distinguish $h$ from the classifier that aligns perfectly with Alice’s report.

We now prove that our protocol satisfies Theorem~\ref{thm:protocol_realizable}. 

\begin{algorithm2e}[htb]
\caption{Critical Points Protocol for Hypothesis Class $\calH$}
\label{alg:abstract_critical_points_protocol}
\DontPrintSemicolon
\LinesNumbered

\KwIn{Subroutine for computing critical points (\Cref{alg:abstract_critical_points_computation}), Labeled points from Alice}
\KwOut{A subset of points sent to Bob}

Alice sends all points $X$ to Trent\;
Alice reports to Trent a set $X_A^+ \subseteq X$ as positive ($X_A^- = X \setminus X_A^+$ as negative)\;


Trent computes the critical points $C(X_A^+)$ using Algorithm~\ref{alg:abstract_critical_points_computation} with input $X_A^+$\;
Trent sends points $X_A^+ \cup C(X_A^+)$ to Bob\;
Bob labels the points and sends labels to Trent\;
Trent checks the agreement of reports from Alice and Bob and sends any disputed points to the court to settle\;

\If{the court disagrees with Alice's label on any disputed points}{
    Trent sends all points $X$ to Bob\;
}

\end{algorithm2e}

\begin{algorithm2e}[htb]
\caption{Computing Critical Points for Hypothesis Class $\calH$}
\label{alg:abstract_critical_points_computation}
\DontPrintSemicolon
\LinesNumbered

\KwIn{A set of $n$ points $X = \{x_1,x_2,\dots, x_n\} \subset \mathbb{R}^d$, a set $X_A^+ \subseteq X$ of positive points reported by Alice, and an oracle $\calO$ for checking realizability within hypothesis class $H$.}
\KwOut{A set of critical points $C(X_A^+) \subseteq X_A^-$, where $X_A^- = X \setminus X_A^+$.}

Set $M = X_A^-$\;
\For{each $x_i \in X_A^-$}{
    Set $T_1 = X_A^+ \cup \{x_i\}$ and $T_2 = M \setminus \{x_i\}$\;
    \If{labeling $T_1$ as $+$ and $T_2$ as $-$ is an {invalid} labeling under $\calH$ according to $\calO$}{
        Remove $x_i$ from $M$, ie, $M = M \setminus \{x_i\}$\;
    }
}
Set the critical points $C(X_A^+) = M$\;

\end{algorithm2e}

\begin{proof}[Proof of Theorem~\ref{thm:protocol_realizable}]
    \textbf{$\mathrm{Recall} = 1$:} Let $X_A^+$ and $X_A^-$ be the set of all positives and the set of all negatives reported by Alice, respectively. 
    Without loss of generality, we assume that all points in $X_A^+$ are true positive. 
    (If this is false, Bob will identify any misclassified negative points within $X_A^+$, as these points are always sent to him for verification.)
    Then, let $\calH(X_A^+)$ be the set of all classifiers in $\calH$ that satisfy the following conditions: (1) all points in $X_A^+$ are still classified as positive; and (2) at least one point in $X_A^-$ is classified as positive. 
    
    We first show that using $C(X_A^+)$ we can distinguish two cases: (1) the labels reported by Alice are correct; and (2) Alice labels some true positive points as negative, i.e. the true classifier is in $\calH(X_A^+)$. 
    Specifically, we show that for any classifier $h \in \calH(X_A^+)$, there exists a critical point in $C(X_A^+)$ classified as positive by $h$. 
    Note that critical points $C(X_A^+) \subseteq X_A^-$. 
    If all critical points $C(X_A^+)$ are true negative, then we are in case (1); otherwise, we are in case (2).
    
    We show this by contradiction. Suppose there is no point in $C(X_A^+)$ classified as positive by $h$. 
    Consider any classifier $h \in \calH(X_A^+)$ that classifies at least a point in $X_A^-$ as positive. We use $X_h^+ = \{x \in X: h(x) = 1\}$ to denote the points classified as positive by $h$. Then, we have $X_h^+ \cap X_A^- \neq \varnothing$. 
    Since no point in $C(X_A^+)$ is classified as positive by $h$, all points in $X_h^+ \cap X_A^-$ are removed in Algorithm~\ref{alg:abstract_critical_points_computation}. 
    Now, consider the last point $x_i$ in $X_h^+ \cap X_A^-$ that is removed in Algorithm~\ref{alg:abstract_critical_points_computation}. Let $M_i$ be the set $M$ at the beginning of the iteration at point $x_i$. Since $x_i$ is the last point in $X_h^+ \cap X_A^-$, we have $M_i \setminus \{x_i\} \cap X_h^+ = \varnothing$. Thus, $T_1 = X_A^+ \cup \{x_i\}$ and $T_2 = M_i \setminus \{x_i\}$ are perfectly classified by the classifier $h \in \calH$, which implies that $x_i$ is not removed. 

    Hence, the protocol always guarantees perfect recall, $\mathrm{Recall} = 1$ since if Alice hides any true responsive documents, then Bob will detect such a document and then the court or Trent will reveal all documents to Bob. 


    \textbf{$\mathrm{Nonresponsive~Disclosure} \leq k$:} We now show that the number of critical points is at most the Leave-One-Out dimension $k$.
    Suppose $C(X_A^+)$ contains $m$ points, denoted as $v_1, v_2, \dots, v_m \in X$. We claim that for each $v_i$, there exists a classifier $h_i \in \calH$ such that $h_i(v_i) = 1$ and $h_i(v_j) = -1$ for all $j \neq i$ (in fact, $h_i \in \calH(X_A^+)$). 
    Consider the iteration corresponding to the point $v_i$ in Algorithm~\ref{alg:abstract_critical_points_computation}. Let $M_i$ be the set $M$ at the beginning of this iteration. Since $v_i$ is not removed, there exists a classifier $h \in \calH$ that separates $\{v_i\}$ from $M_i \setminus \{v_i\}$. Noting that the critical points $C(X_A^+)$ form a subset of $M_i$, this classifier $h$ satisfies the desired property.
    Thus, for each point $v_i$ in $C(X_A^+)$, there exists a classifier in $\calH$ that classifies $v_i$ as positive while classifying all other points in $C(X_A^+)$ as negative. By Definition~\ref{defn:leave_one_out}, it follows that the number of critical points is at most the Leave-One-Out dimension of $\calH$, i.e., $|C(X_A^+)| \leq k$.

    Overall, if Alice labels all documents of $X$ correctly, then the protocol only discloses nonresponsive documents in $C(X_A^+)$ to Bob, which is at most $k$ in size.

    \emph{Truthfulness and Runtime:} Since reporting false labels is guaranteed to reveal all the documents to Bob, it is in Alice's best interest to be truthful. 
    Further, as we only call the membership oracle $\calO$ one per document in $X$ the runtime is bounded by $O(|X|)$.
    
    \emph{Lower Bound:} Let $C \subseteq X$ be the subset witnessing the Leave-One-Out dimension of $X$, so that $|C| = k$ and for each $c \in C$, there exists a classifier $h_c \in \calH$ satisfying $h_c(c) = 1$ and $h_c(c') \neq 1$ for all $c' \in C \setminus \{c\}$. 
    Choose an arbitrary $c \in C$ with its corresponding hypothesis $h_c$, and define $X' = C \setminus \{c\}$. Further, set $f(x) = -1$ for all $x \in X'$. Clearly, $(X', f)$ is realized by $h_c$ when restricted to $X'$. 
    Now, consider any protocol with recall $1$. If the protocol discloses fewer than $k-1$ nonresponsive documents, then there exists some $c' \in X'$ that is not revealed. However, the ground truth could instead be given by $h_{c'}$, which labels $c'$ as $+1$ and all other points in $X'$ as $-1$. In this case, the protocol fails to reveal all relevant documents to Bob, leading to a contradiction.

\end{proof}

\subsection{Leave-One-Out Dimension of Linear Classifiers with Margin}\label{sec:realizable_linear_margin}

In this section, we characterize the Leave-One-Out dimension of the linear classifiers with a margin.

Without loss of generality, we assume that the linear classifier passes through the origin in $\R^d$, as one can add an extra dimension for the bias if needed. Let $w \in \R^d$ be the unit-length weight vector of a linear classifier $h$, so that for any $x \in X$, we have $h(x) = \operatorname{sign}(\langle w, x \rangle)$.

We define the margin of a linear classifier as follows. 
\begin{definition}
\label{defn:margin}
    Given a set of points $X \subset \R^d$, the \emph{margin} of a linear classifier $h$ with unit-length weight vector $w$ on $X$ is defined as 
    $$
    \gamma(w,X) = \min_{x \in X} \frac{|\inner{w}{x}|}{\|x\|_2}.
    $$
\end{definition}
Then, for any $\gamma \in [0,1]$ and $X\subseteq \R^d$, we use $\calH_\gamma(X)$ to denote the hypothesis class of all linear classifiers with a margin at least $\gamma$ on $X$.

We show the following trichotomy of the Leave-One-Out dimension of the hypothesis class $\calH_\gamma(X)$.

\begin{theorem}\label{thm:Leave-One-Out-linear-realizable}
    For any set of points $X \subset \R^d$, any $\gamma \in [0,1]$, the Leave-One-Out dimension $k$, of the class $\calH_\gamma(X)$ is 
    \begin{enumerate}
        \item $k \leq \frac{2+2\gamma}{3\gamma -1}$,  if $\gamma > \frac{1}{3}$.
    \end{enumerate}
    Further, there are sets $X$ such that 
    \begin{enumerate}
        \item $k \geq \Omega(d)$, if $\gamma = \frac{1}{3}$;
        \item $k \geq \exp(\Omega((1/3 - \gamma)^2d))$, if $\gamma < \frac{1}{3}$.
    \end{enumerate}
    
\end{theorem}

\begin{corollary}
    Armed with Theorems~\ref{thm:protocol_realizable} and \ref{thm:Leave-One-Out-linear-realizable}, we analyze instances that are linearly separable with a large margin $\gamma > 1/3$, i.e., there exists $w \in \mathbb{R}^d$ such that for all $x \in X$:
\[
\frac{\langle x, w \rangle \cdot f(x)}{\|x\|_2} \geq \gamma.
\]
For such instances, \Cref{alg:abstract_critical_points_protocol} achieves a nonresponsive disclosure of at most $O(1)$, independent of the instance size $|X|$ and ambient dimension $d$.  
Furthermore, an efficient oracle $\calO$ verifies label membership in this class using the hard support vector machine (SVM) (see Theorem 15.8 in~\citet*{shalev2014understanding}). Thus, \Cref{alg:abstract_critical_points_protocol} is overall efficient.

For instances with smaller margins, however, any protocol ensuring $\mathrm{Recall} = 1$ must incur a nonresponsive disclosure that scales either linearly with $d$ when $\gamma = 1/3$, or exponentially with $d$ when $\gamma < 1/3$.

\end{corollary}
The proof of \Cref{thm:Leave-One-Out-linear-realizable} follows directly from \Cref{lemma:skew-obtuse_lemma} by interpreting $V$ and $W$ in the lemma as $C$ (the set witnessing the \emph{Leave-One-Out dimension}) and the corresponding hypothesis $h_c$, respectively.
The lemma concerns the geometry of a \emph{skew-obtuse family of vectors}, a result that may be of \emph{independent interest} with applications to \emph{coding theory} and \emph{combinatorial geometry}. Similar statements about standard obtuse vector families are used to establish distance bounds.  
Additionally, we believe that the proof is more {nuanced} than the usual approach of analyzing the \emph{Gram matrix} of the vector family.
The proof is detailed in \Cref{apx:constructing_skob_family}

\begin{lemma}[Skew-Obtuse Family of Vectors]
\label{lemma:skew-obtuse_lemma}
    Let $V= \bar{v}_1,\bar{v}_2,\cdots, \bar{v}_k$ be a $k$ unit vectors in $\R^d$. Further, let  $W = w_1,w_2,\cdots, w_k$ be $k$ other unit vectors in $\R^d$. Suppose there exists a $\gamma\in [0,1]$ such that the following holds:
    \begin{enumerate}
        \item $\inner{\bar{v}_i}{w_i} \geq \gamma$ and
        \item $\inner{\bar{v}_i}{w_j} \leq -\gamma$ if $i\neq j$.
    \end{enumerate}
     Then, if $\gamma>1/3$ we have $k \leq \frac{2+2\gamma}{3\gamma-1}.$ Further if $\gamma=1/3$, there exists a skew-obtuse family of vectors with $k\geq \Omega(d)$ and if $\gamma<1/3$ then there exists such a family with $k\geq \exp{(\Omega(({1/3-\gamma})^2d))}$.
\end{lemma}

\section{Nonrealizable Setting}
\label{sec:non_realizable}

In this setting, we consider instances that can be classified by a classifier $h$ in a hypothesis class $\calH$, not necessarily \emph{perfectly}, but with a small number of errors.
We first introduce the \emph{robust Leave-One-Out dimension} and provide a robust protocol for general hypothesis classes that achieves high recall and nonresponsive disclosure at most the robust Leave-One-Out dimension of $\calH$.
Then, we characterize the robust Leave-One-Out dimension of linear classifiers with a margin. 

\begin{definition}[robust Leave-One-Out dimension]
\label{defn:robust_leave_one_out}
     Given a set system $(X,\calS)$ and a slack $L \in \N$, the \emph{robust Leave-One-Out} dimension is the cardinality of the largest set $C\subseteq X$ such that for each element $c \in C$, there exists $S \in \calS$ with $c \in S \cap C$ and $|S\cap C| \leq L+1$.
\end{definition}

\begin{theorem}\label{thm:robust_protocol_general}
    Let $\calH$ be a hypothesis class of binary classifiers on a set $X$ with robust Leave-One-Out dimension $k$ when the slack is $L$.  
    Suppose $f: X \to \{+1,-1\}$ represents the true labels for responsive and nonresponsive documents, and let $(X,f)$ be an instance classified by a classifier in $\calH$ with at most $L$ errors. Then, \Cref{alg:robust_abstract_critical_points_protocol} (in Appendix~\ref{apx:robust_protocol}) defines a multi-party verification protocol with the following properties:
    \begin{enumerate}
        \item \textbf{(Recall)} The recall is at least $1-L/n^+$, where $n^+$ is the cardinality of the true responsive documents.
        \item \textbf{(Nonresponsive Disclosure)} If Alice reports all labels correctly, the number of disclosed nonresponsive documents is at most $k$.
        \item \textbf{(Efficiency)} Given an empirical risk minimization (ERM) oracle $\calO$ for $\calH$ (see \Cref{asp:oracle}), the protocol runs in time $O(|X|)$.
    \end{enumerate}
\end{theorem}


The proof of \Cref{thm:robust_protocol_general} appears in Appendix~\ref{apx:robust_protocol} where we also describe \Cref{alg:robust_abstract_critical_points_protocol} in detail. 
The protocol follows the same framework as Algorithm~\ref{alg:abstract_critical_points_protocol}, but instead of using Algorithm~\ref{alg:abstract_critical_points_computation}, it uses Algorithm~\ref{alg:robust_critical} as a subroutine to compute robust critical points.
We also assume access to the following empirical risk minimization (ERM) oracle.

\begin{assumption}\label{asp:oracle}
    Suppose there is an oracle $\calO$ that given any instance $(X,f)$, a hypothesis class $\calH$, a point $x \in X$, finds a classifier in $\calH$ that classifies $x$ as positive and minimizes the total error among all classifiers in $\calH$ that label $x$ as positive.
\end{assumption}

\begin{algorithm2e}[htb]
\caption{Computing Robust Critical Points}
\label{alg:robust_critical}
\DontPrintSemicolon
\LinesNumbered

\KwIn{A set of $n$ points $X = \{x_1,x_2,\dots, x_n\} \subset \mathbb{R}^d$, a set $X_A^+ \subseteq X$ of positive points reported by Alice, an error tolerance $L \in \mathbb{N}$, and an oracle $\calO$ as in Assumption~\ref{asp:oracle}.}
\KwOut{A set of critical points $C_{L}(X_A^+) \subseteq X_A^-$, where $X_A^- = X \setminus X_A^+$.}

Set $M = X_A^-$\;
\For {each $x_i \in X_A^-$}{
    Set $T_1 = X_A^+ \cup \{x_i\}$ and $T_2 = M \setminus \{x_i\}$\;
    Create an instance $(X',f')$ with $X'=T_1 \cup T_2$ labeling $T_1$ as positive and $T_2$ as negative\;
    Find a classifier in $\calH$ with error at most $L$ on $(X',f')$ that classifies $x_i$ as positive using the oracle $\calO$\;
    \If{no such classifier in $\calH$}{
        Remove $x_i$ from $M$, $M = M \setminus \{x_i\}$\;
    }
}
Set the critical points $C_{L}(X_A^+) = M$\;

\end{algorithm2e}






We now characterize the robust Leave-One-Out dimension of linear classifiers with a fixed margin. 
In the nonrealizable setting, a large-margin classifier may incur two types of errors: (1) margin error; and (2) classification error. The margin errors occur when points are classified correctly but do not satisfy the margin condition. 
The classification errors occur when points are misclassified by the classifier. 
We consider instances that can be classified by a large margin classifier with a small total error, where the total error contains both the margin errors and the classification errors. 

\begin{definition}
    An instance $(X,f)$ is classified by a linear classifier with margin $\gamma \in [0,1]$ and total error $L \in \N$ if the number of total error is at most
    $$
    |\{x: \inner{\bar{x}}{w} \cdot f(x) < \gamma\}|  \leq L,
    $$
    where $w$ is the weight vector of this classifier and $\bar{x} = x / \|x\|$ be the normalized vector for point $x$.
\end{definition}

For any instance $(X,f)$, any margin $\gamma \in [0,1]$, and any error slack $L \in \N$, let $\calH_{\gamma,L}(X,f)$ be the class of linear classifiers that classifies $(X,f)$ with margin $\gamma$ and total error at most $L$. We then characterize the robust Leave-One-Out dimension of this class $\calH_{\gamma,L}(X,f)$ for a large margin $\gamma > 1/3$. 

\begin{theorem}\label{thm:robust_leave-one-out}
    For any instance $(X,f)$, any $1/3 < \gamma \leq 1$, and any error slack $L \in \N$, the robust Leave-One-Out dimension of the class $\calH_{\gamma, L}(X,f)$ is 
    $$
        \frac{(2+2L)(\gamma+1)}{3\gamma -1}.
    $$
\end{theorem}

The proof of Theorem~\ref{thm:robust_leave-one-out} follows from a generalized version of Lemma~\ref{lemma:skew-obtuse_lemma}, which extends the geometric properties of a \emph{skew-obtuse family of vectors}. The argument mirrors the way \Cref{thm:Leave-One-Out-linear-realizable} follows from \Cref{lemma:skew-obtuse_lemma}.  
We apply \Cref{lemma:generalized-skew-obtuse_lemma} with $\alpha = \beta = \gamma$. Let $C \subseteq X$ be a set witnessing the robust Leave-One-Out dimension for $\calH_{\gamma,L}(X,f)$. For any $c \in C$, there exists a hypothesis $h_c \in \calH_{\gamma,L}(X,f)$ that classifies $c$ as positive and at most $L$ other points in $C$ as positive. Additionally, $h_c$ may classify some points in $C$ as negative but with low margin.  
By definition of $\calH_{\gamma,L}(X,f)$, at most $L$ points are either misclassified or have low margin. Thus, for each $c \in C$, there exists a unit vector $w_c$ (the weight vector of $h_c$) satisfying the two properties in \Cref{lemma:generalized-skew-obtuse_lemma}.  
The proof of this lemma is deferred to Appendix~\ref{apx:proof_general_skew_obtuse}.

\begin{lemma}[Robust Skew-Obtuse Family of Vectors]
\label{lemma:generalized-skew-obtuse_lemma}
    Let $V= \bar{v}_1,\bar{v}_2,\cdots, \bar{v}_k$ be a $k$ unit vectors in $\R^d$. Further, let  $W = w_1,w_2,\cdots, w_k$ be $k$ other unit vectors in $\R^d$. Suppose there exists $\alpha,\beta\in [0,1]$ and an error parameter $L$ such that the following holds:
    \begin{enumerate}
        \item $\inner{\bar{v}_i}{w_i} \geq \alpha$ and
        \item for all $j\in [k]$ we have $\left|i\in[k]\colon  \inner{\bar{v}_i}{w_j} \leq -\beta\right|\geq k-1-L$. 
    \end{enumerate}
     Then, if $\alpha+2\beta>1$ we have $ k \leq \frac{(2+2L)(1+\beta)}{\alpha+2\beta-1}.$
\end{lemma}

\section{Error-Tolerant Protocol}

In this section, we provide protocols that are tolerant of classification errors made by Alice. For protocols proposed in previous sections, if Bob detects one document misclassified by Alice, the court or Trent reveals all documents to Bob. 
However, \citet*{grossman2010technology,grossman2012inconsistent} show that, in practice, even the most skilled human reviewers can unintentionally make classification errors.
Therefore, we aim to design protocols that are less strict on Alice's classification, imposing a gradual rather than harsh penalty for Alice's misclassification errors.
We show that our previous protocols can be converted to the following error-tolerant protocols.

\begin{theorem}
\label{thm:error_tolerant_protocol}
    Suppose there exists a multi-party classification protocol that satisfies if Bob does not detect any document misclassified by Alice, then the recall is at least $\alpha$ and the nonresponsive disclosure is at most $k$.
    Then, there exists an error-tolerant protocol that guarantees:
    \begin{enumerate}
        \item \textbf{(Recall)} The recall is at least $\alpha$;
        \item \textbf{(Nonresponsive Disclosure)} The nonresponsive disclosure is at most $k\cdot E$ if the protocol detects $E$ documents misclassified by Alice.
    \end{enumerate}
\end{theorem}

\begin{proof}
    The error-tolerant protocol is constructed as follows. Let $P$ be a multi-party classification protocol that ensures a recall of at least $\alpha$ and discloses at most $k$ nonresponsive documents when Bob does not detect any document misclassified by Alice. (This is not the number of errors made by the optimal classifier.)
    The error-tolerant protocol iteratively calls protocol $P$ until an iteration of $P$ completes without any detected misclassification by Alice.
    
    If Bob identifies a document misclassified by Alice in any call of $P$, Alice is required to relabel the document correctly. 
    Since each detected misclassification triggers at most $k$ additional nonresponsive disclosures per iteration, the total nonresponsive disclosure remains bounded by $k \cdot E$, where $E$ is the number of errors detected.
\end{proof}

\begin{algorithm2e}[htb]
\caption{Computing Critical Points with Verified Points}
\label{alg:critical_negative}
\DontPrintSemicolon
\LinesNumbered

\KwIn{A set of $n$ points $X = \{x_1,x_2,\dots, x_n\} \subset \mathbb{R}^d$, a set $A^- \subseteq X$ of verified negative points, a set $X_A^+ \subseteq X$ of positive points reported by Alice, and an oracle $\calO$ for checking realizability within hypothesis class $H$.}
\KwOut{A set of critical points $C(X_A^+, A^-) \subseteq X_A^-$, where $X_A^- = X \setminus X_A^+$.}

Set $M = X_A^-$\;
\For{each $x_i \in X_A^- \setminus A^-$}{
    Set $T_1 = X_A^+ \cup \{x_i\}$ and $T_2 = M \setminus \{x_i\}$\;
    \If{labeling $T_1$ as $+$ and $T_2$ as $-$ is an {invalid} labeling under $\calH$ according to $\calO$}{
        Remove $x_i$ from $M$, ie, $M = M \setminus \{x_i\}$\;
    }
}
Set the critical points $C(X_A^+,A^-) = M$\;

\end{algorithm2e}

Note that the error-tolerant protocol framework shown in Theorem~\ref{thm:error_tolerant_protocol} may call the one-round multi-party protocol multiple times. 
If Bob detects misclassified documents, then in later calls, instead of certified positive documents reported by Alice, there may also be some negative documents verified by Bob in previous calls. 
Thus, we can modify the algorithms for computing the critical points (Algorithm~\ref{alg:abstract_critical_points_computation}) and robust critical points (Algorithm~\ref{alg:robust_critical}) to achieve fewer nonresponsive disclosure by taking those certified negative documents into account. 
We show the modified algorithm for computing the critical points in the realizable setting in Algorithm~\ref{alg:critical_negative}.
Algorithm~\ref{alg:robust_critical} for the robust critical points in the nonrealizable setting can be extended similarly.
In practice, using these modified algorithms to compute the critical points in the error-tolerant protocol can further reduce the nonresponsive disclosure while preserving the recall guarantee.
\section{Conclusion}

We presented a multi-party classification protocol that verifies a hypothesized labeling with limited disclosure of nonresponsive documents. Its key insight is that the number of nonresponsive points disclosed is fundamentally tied to the Leave-One-Out dimension of the hypothesis class. For linear classifiers with large margins, the bound on nonresponsive disclosure is independent of the dimension, providing a strong privacy guarantee. We also extended the protocol to nonrealizable settings by introducing a robust Leave-One-Out dimension that accounts for classification errors, offering similarly efficient verification with minimal disclosure.

A few immediate avenues remain open. First, our skew-obtuse family analysis could be adapted from strict margins to average-margin conditions. Thus, it would be interesting to generalize our protocols for verifying classifiers with an “expected” large margin. 
Second, it would be worthwhile to investigate whether additional structural properties of the hypothesis class—beyond margin separation—can be exploited for efficient verification, especially in the nonrealizable case.

\section*{Acknowledgments}

We thank Jinshuo Dong, Jason Hartline, and Aravindan Vijayaraghavan for helpful discussions. 


\bibliographystyle{plainnat}
\bibliography{ref}

\newpage

\appendix

\section{Proof of Lemma~\ref{lemma:skew-obtuse_lemma}}\label{apx:constructing_skob_family}


\begin{proof}
    Let $V= (\bar{v}_1,\bar{v}_2,\cdots, \bar{v}_k)^T$ be a $k\times d$ matrix. 
    Similarly, we define $W = (w_1,w_2,\cdots, w_k)^T$ be a $k\times d$ matrix. Then, we have $VW^T$ is a $k\times k$ matrix, where each entry is $\inner{\bar{v}_i}{w_j}$. 
    Let $\{a_1, a_2, \dots, a_d\}$ be the $d$ column vectors of matrix $V$. Let $\{b_1,b_2,\dots, b_d\}$ be the $d$ column vectors of $W$.

    We first consider the trace of matrix $VW^T$. Since each diagonal entry of $VW^T$ satisfies $\inner{\bar{v}_i}{w_j} \geq \gamma$, we have 
    $$
    \Tr(VW^T) = \sum_{i=1}^k \inner{\bar{v}_i}{w_i} \geq k \gamma.
    $$
    Since $\bar{v}_i$ and $w_i$ are all unit vectors, we have $\Tr(VW^T) \leq k$.
    We also have 
    $$
    \Tr(VW^T) = \Tr(WV^T) = \sum_{i=1}^d \inner{a_i}{b_i} = \sum_{i=1}^d \|a_i\|\|b_i\|\cos(\theta_i),
    $$
    where $\theta_i$ is the angle between vectors $a_i$ and $b_i$. Hence, we get
    \begin{equation}\label{eq:trace}
    \sum_{i=1}^d \|a_i\|\|b_i\|\cos(\theta_i) = \Tr(VW^T) \geq k\gamma.
    \end{equation}

    We now consider the sum of all entries in matrix $VW^T$. Since all off-diagonal entries of matrix $VW^T$ satisfies $\inner{\bar{v}_i}{w_j} \leq -\gamma$. We use $\one$ to denote all one vector. Then, we have the sum of all entries is
    $$
    \one^T VW^T \one = \Tr(VW^T) + \sum_{i\neq j} \inner{\bar{v}_i}{w_j} \leq \Tr(VW^T) - k(k-1)\gamma.
    $$
    Note that $VW^T = \sum_{i=1}^d a_i b_i^T$. Thus, we have $\one^T VW^T \one = \sum_{i=1}^d \inner{\one}{a_i} \inner{b_i}{\one}$. Let $\alpha_i$ be the angle between $\one$ and $-a_i$ and $\beta_i$ be the angle between $\one$ and $b_i$. Then, we have 
    $$
    - \one^T VW^T \one = \sum_{i=1}^d \inner{\one}{-a_i} \inner{b_i}{\one} = \sum_{i=1}^d \|\one\|^2 \|a_i\|\|b_i\| \cos(\alpha_i) \cos(\beta_i). 
    $$
    Since the angle between $-a_i$ and $b_i$ is $\pi - \theta_i$, we have $\alpha_i + \beta_i \geq \pi - \theta_i$. Since the cosine function is decreasing and log-concave on $[0,\pi/2]$, and negative in $[\pi/2, \pi]$, we have 
    $$
    \cos(\alpha_i) \cos(\beta_i) \leq \cos^2\left(\frac{\pi-\theta_i}{2}\right).
    $$
    By combining the equations above, we have
    \begin{equation}\label{eq:all_sum}
    k \sum_{i=1}^d \|a_i\|\|b_i\| \cos^2\left(\frac{\pi-\theta_i}{2}\right) \geq - \one^T VW^T \one \geq - \Tr(VW^T) + k(k-1)\gamma \geq - k + k(k-1)\gamma,
    \end{equation}
    where the last inequality is due to $\Tr(VW^T) \leq k$.

    By combining Equations~(\ref{eq:trace}) and~(\ref{eq:all_sum}), we have 
    $$
    \sum_{i=1}^d \|a_i\|\|b_i\| = \sum_{i=1}^d \|a_i\|\|b_i\| \left(2\cos^2\left(\frac{\pi-\theta_i}{2}\right) + \cos(\theta_i)\right) \geq 3k\gamma -2 - 2\gamma.
    $$
    By Cauchy-Schwarz inequality, we have 
    $$
    \left(\sum_{i=1}^d \|a_i\|\|b_i\| \right)^2 \leq \sum_{i=1}^d \|a_i\|^2 \cdot \sum_{i=1}^d \|b_i\|^2 = k^2,
    $$
    where the last equality is from $\sum_{i=1}^d \|a_i\|^2 = \sum_{i=1}^d \|b_i\|^2 = k$ since matrices $V$ and $W$ are both consists of $k$ unit row vectors. Therefore, we have 
    $$
    k \leq \frac{2+2\gamma}{3\gamma-1}.
    $$

    For the lower bounds see below.
\end{proof}

\begin{proof}[Construction of skew-obtuse family of vectors]
     Let $\varepsilon = 1/\sqrt{3}$.
    We first consider the case when $\gamma =1/3$.
   The dataset $X$ contains $n$ data points $x_1,\dots,x_n$ in $\bbR^{n+1}$ such that 
    $$
    x_i = \varepsilon \cdot e_1 + \sqrt{1-\varepsilon^2} \cdot e_{i+1},
    $$
    where $e_i$ is the $i$-th standard basis vector. Each data point $x_i$ has a unit length. 

    Consider the following $n$ different label functions $f_1,\dots, f_n$ on this dataset $X$. The function $f_i$ labels the point $x_i$ as positive and all other points as negative. 
    Let $h_i$ be the linear classifier with the weight vector $w_i = -\varepsilon \cdot e_1 + \sqrt{1-\varepsilon^2} \cdot e_{i+1}$. Then, we have $\inner{h_i}{x_i} = 1-2\varepsilon^2  \geq \gamma$ and $\inner{h_i}{x_j} = -\varepsilon^2  \leq -\gamma$ for all $j \neq i$. 
    Thus, each instance $(X,f_i)$ is linear separable by margin $\gamma$ and $h_i$ is a $\gamma$-margin classifier for this instance $(X,f_i)$. Therefore, to distinguish these instances and achieve recall $1$, any protocol must reveal all data points for verification.

    Next we consider the case when $\gamma = 1/3 - \eta$ for some $\eta >0$.
    Let $\beta = (3/2)\eta$.
    It is well-known that by picking unit vectors randomly in $\R^d$ we can construct a collection of vectors $v_1, v_2, \ldots v_k$ such that $\inner{v_i}{v_j} \leq \beta$ if $i\neq j$ and $k\geq \exp{(\Omega(\beta^2 d))}$ (see \Cref{lem:exp_many_neearly_orthogoanl_vectors} below for a proof). 
    WLOG we can assume that $v_1,\ldots,v_k$ are in the span of $e_2,\ldots,e_{d+1}$. Now, for all $i\in [k]$ we set 
    $$
    x_i = \varepsilon \cdot e_1 + \sqrt{1-\varepsilon^2} \cdot v_{i}.
    $$
    Further, by setting $w_i = -\varepsilon \cdot e_1 + \sqrt{1-\varepsilon^2} \cdot v_i$ we have that 
    \[
    \inner{v_i}{w_i} = -\varepsilon^2 + 1 - \varepsilon^2 = 1/3 \geq \gamma
    \]
    and for $i\neq j$
    \[
    \inner{v_i}{w_j} \leq -\varepsilon^2 + (1 - \varepsilon^2)\eta = -1/3 + \eta = -\gamma.
    \]
\end{proof}

\begin{lemma}[Exponentially many nearly orthogonal vectors]
\label{lem:exp_many_neearly_orthogoanl_vectors}
For any $\epsilon > 0$, there exists a collection of $N = \exp(c\epsilon^2d)$ unit vectors in $\mathbb{R}^d$ such that the absolute value of the inner product between any pair is at most $\epsilon$, where $c > 0$ is an absolute constant.
\end{lemma}

\begin{proof}
Let us construct the vectors by selecting $N = \exp(c\epsilon^2d)$ vectors independently and uniformly at random from $\{\pm \frac{1}{\sqrt{d}}\}^d$. We will show that with positive probability, all pairwise inner products are bounded by $\epsilon$ in absolute value.

For any fixed pair of vectors $v_1, v_2$ chosen according to this distribution we an analyze the inner product $\langle v_1, v_2 \rangle$ as the sum of $d$ independent random variables where each term in this sum has mean zero and magnitude $\frac{1}{d}$.

By Hoeffding's inequality, for any fixed pair of vectors:
\[
    \mathbb{P}(|\langle v_1, v_2 \rangle| > \epsilon) \leq 2\exp(-2\epsilon^2d)
\]

By the union bound:
\[
    \mathbb{P}(\exists i,j: |\langle v_i, v_j \rangle| > \epsilon) \leq \binom{N}{2} \cdot 2\exp(-2\epsilon^2d) < N^2\exp(-2\epsilon^2d)
\]

Substituting $N = \exp(c\epsilon^2d)$ where $c < 1$:
\[
    N^2\exp(-2\epsilon^2d) = \exp(2c\epsilon^2d - 2\epsilon^2d) < 1
\]

Therefore, with positive probability, all pairwise inner products are at most $\epsilon$ in absolute value, proving the existence of such a collection.
\end{proof}

\section{Proof of Theorem~\ref{thm:robust_protocol_general}}\label{apx:robust_protocol}

In this section, we prove the guarantees of our robust critical points protocol given below.

\begin{algorithm2e}[htb]
\caption{Robust Critical Points Protocol for Hypothesis Class $\calH$}
\label{alg:robust_abstract_critical_points_protocol}
\DontPrintSemicolon
\LinesNumbered

\KwIn{Subroutine for computing robust critical points (\Cref{alg:robust_critical}), Labeled points from Alice}
\KwOut{A subset of points sent to Bob}

Alice sends all points $X$ to Trent\;
Alice reports to Trent a set $X_A^+ \subseteq X$ as positive ($X_A^- = X \setminus X_A^+$ as negative)\;


Trent computes the critical points $C(X_A^+)$ using Algorithm~\ref{alg:robust_critical} with input $X_A^+$\;
Trent sends points $X_A^+ \cup C(X_A^+)$ to Bob\;
Bob labels the points and sends labels to Trent\;
Trent checks the agreement of reports from Alice and Bob and sends any disputed points to the court to settle\;

\If{the court disagrees with Alice's label on any disputed points}{
    Trent sends all points $X$ to Bob\;
}

\end{algorithm2e}

\begin{proof}[Proof of \Cref{thm:robust_protocol_general}]
    Let $h^*$ be the best classifier in the hypothesis class $\calH$ on the true labels $(X,f)$. We know that $h^*$ classifies $(X,f)$ with at most $L$ errors.
    Let $X_A^+$ and $X_A^-$ be the set of all positives and the set of all negatives reported by Alice, respectively. Let $f_A$ be this labeling function reported by Alice.
    Without loss of generality, we assume that all points in $X_A^+$ are true positive. 
    (If this assumption is false, Bob will identify any misclassified negative points within $X_A^+$, as these points are always sent to him for verification.)
    Next, define $\calH_{L}(X_A^+)$ be the set of all classifiers in $\calH$ that satisfies the following conditions: (a) the instance $(X,f_A)$ are not classified by this classifier with error at most $L$; and (b)
    there is a labeling $f'$ of $X$ such that all points in $X_A^+$ are labeled as positive, i.e.,  $f'(x) = +1$ for all $x\in X_A^+$ and the labeled dataset $(X,f')$ is classified by this classifier with error at most $L$. 
    Without ambiguity, we use $\calH_L$ to denote this hypothesis set. 
    
    We first show that the critical points $C_{L}(X_A^+)$ computed by Algorithm~\ref{alg:robust_critical} can distinguish two cases: (1) Alice's report $(X,f_A)$ is classified by the best classifier $h^*$ with error at most $L$; and (2) Alice labels some true positive points as negative. 
    Specifically, we show that there exists a true positive point in  $C_{L}(X_A^+)$ if the best classifier $h^*$ is in $\calH_L$. If there is a true positive in $C_{L}(X_A^+)$, then we are in case (2); otherwise, the best classifier $h^*$ is not in $\calH_L$, which means we are in case (1) since the condition (b) of $\calH_L$ is always satisfied by the best classifier $h^*$.
    
    Suppose the best classifier $h^*$ is $h \in \calH_L$. Let $E^- = \{x \in X_A^+: h(x) = -1 \}$ be the points in $X_A^+$ that are classified by $h$ as negative. Since $h \in \calH_L$, by the definition of $\calH_L$, we have $|E^-| \leq L$ since errors in $E^-$ can not be avoided by relabeling $X_A^-$.
    Let $E^+ = \{x \in X_A^-: h(x) = +1\}$ be the points in $X_A^-$ that are classified by $h$ as positive. We must have $|E^+| > L-|E^-| \geq 0$ since if $|E^+| \leq L-|E^-|$, then $(X,f_A)$ is classified by $h$ with error at most $L$, which contradicts $h \in \calH_L$. 
    Consider the last $L-|E^-|+1$ points in $E^+$ visited in Algorithm~\ref{alg:robust_critical}, denoted by $C_h$.
    When Algorithm~\ref{alg:robust_critical} visits these points, by flipping the label of the visited point, the classifier $h$ can classify the new instance with error at most $L$. Thus, all these $L-|E^-|+1$ points $C_h$ are not removed and are contained in critical points $C_{L}(X_A^+)$. 
    Since $h$ is the best classifier, there exists at least a true positive in these $L-|E^-|+1$ points $C_h$, otherwise, $h$ makes more than $L$ errors on the true labels. Thus, this implies there exists a true positive point in $C_{L}(X_A^+)$.

    We now show that the number of robust critical points in $C_{L}(X_A^+)$ is at most the robust Leave-One-Out dimension of the hypothesis class $\calH$. 
    Suppose $C_{L}(X_A^+)$ contains $k$ points $v_1,v_2, \dots, v_k \in X$. 
    We now show that for each $v_i$, there exists a classifier $h_i \in \calH$ satisfies two properties: (1) $v_i$ is classified as positive, i.e. $h(v_i) = +1$; and (2) there are at least $k-1-L$ points in $\{v_i\}$ are classified as negative, $\left|\{j\in[k]\colon  h(v_j) = -1\}\right|\geq k-1-L$.
    Consider the iteration corresponding to the point $v_i$ in Algorithm~\ref{alg:robust_critical}. Let $M_i$ be the set $M$ at the beginning of this iteration. Since $v_i$ is not removed, there is a classifier $h \in \calH$ that classifies $v_i$ as positive and has at most $L$ errors. Note that the critical points $C_L(X_A^+)$ is a subset of $M_i$. This classifier $h$ satisfies two properties for the point $v_i$.
    Therefore, the number of robust critical points is at most the robust Leave-One-Out dimension of the hypothesis class $\calH$.

    Finally, we bound the recall and nonresponsive disclosure. 
    We first show that the recall is at least $1-L/n^+$. 
    If Bob detects any documents misclassified by Alice, then all documents are disclosed to him, which implies a perfect recall, $\mathrm{Recall} = 1$. 
    If Bob does not detect any misclassified document, then by the above analysis, we are in case (1) Alice's report $(X,f_A)$ is classified by the best classifier with error at most $L$.
    Since points in $X_A^+$ are always sent to Bob, Alice can only hide true positive points as negative.
    We now show that Alice can hide at most $L$ true positives.
    Since the true labels are classified by the best classifier with at most $L$ errors, there are at most $L$ true positive points classified as negative by the best classifier.
    If Alice hides more than $L$ true positives as negative, then there exists a true positive point $x$ that is labeled as negative by Alice and is classified as positive by the best classifier.
    When Algorithm~\ref{alg:robust_critical} visits this point $x$, this point is considered a critical point in $C_L(X_A^+)$ since the best classifier satisfies the condition.
    Thus, Alice can hide at most $L$ true positives as negative.
    If Alice labels all documents correctly, then Bob will not detect any misclassified documents. Thus, the nonresponsive disclosure is the number of robust critical points, which is at most the robust Leave-One-Out dimension of the hypothesis class $\calH$. 
    
\end{proof}

\section{Proof of Lemma~\ref{lemma:generalized-skew-obtuse_lemma}}\label{apx:proof_general_skew_obtuse}

In this section, we prove the size of the robust generalized skew-obtuse family of vectors. 


\begin{proof}[Proof of Lemma~\ref{lemma:generalized-skew-obtuse_lemma}]
    Let $V= (\bar{v}_1,\bar{v}_2,\cdots, \bar{v}_k)^T$ be a $k\times d$ matrix. 
    Similarly, we define $W = (w_1,w_2,\cdots, w_k)^T$ be a $k\times d$ matrix. Then, we have $VW^T$ is a $k\times k$ matrix, where each entry is $\inner{\bar{v}_i}{w_j}$. 
    Let $\{a_1, a_2, \dots, a_d\}$ be the $d$ column vectors of matrix $V$. Let $\{b_1,b_2,\dots, b_d\}$ be the $d$ column vectors of $W$.

    We first consider the trace of matrix $VW^T$. Since each diagonal entry of $VW^T$ satisfies $\inner{\bar{v}_i}{w_j} \geq \alpha$, we have 
    $$
    \Tr(VW^T) = \sum_{i=1}^k \inner{\bar{v}_i}{w_i} \geq k \alpha.
    $$
    Since $\bar{v}_i$ and $w_i$ are all unit vectors, we have $\Tr(VW^T) \leq k$.
    We also have 
    $$
    \Tr(VW^T) = \Tr(WV^T) = \sum_{i=1}^d \inner{a_i}{b_i} = \sum_{i=1}^d \|a_i\|\|b_i\|\cos(\theta_i),
    $$
    where $\theta_i$ is the angle between vectors $a_i$ and $b_i$. Hence, we get
    \begin{equation}\label{eq:robust-trace}
    \sum_{i=1}^d \|a_i\|\|b_i\|\cos(\theta_i) = \Tr(VW^T) \geq k\alpha.
    \end{equation}

    We now consider the sum of all entries in matrix $VW^T$. 
    For each column $i \in [k]$ of the matrix $VW^T$, we know that there are at least $k-1-L$ off-diagonal entries with value at most $\inner{\bar{v}_j}{w_i} \leq -\beta$.
    For other off-diagonal entries in that column, we upper bound them by one since $\bar{v}_j$ and $w_i$ are unit vectors.
    We use $\one$ to denote all one vector. Then, we have the sum of all entries is
    $$
    \one^T VW^T \one = \Tr(VW^T) + \sum_{i\neq j} \inner{\bar{v}_j}{w_i} \leq \Tr(VW^T) - k(k-1 -L)\beta + kL.
    $$
    Note that $VW^T = \sum_{i=1}^d a_i b_i^T$. Thus, we have $\one^T VW^T \one = \sum_{i=1}^d \inner{\one}{a_i} \inner{b_i}{\one}$. Let $\phi_i$ be the angle between $\one$ and $-a_i$ and $\psi_i$ be the angle between $\one$ and $b_i$. Then, we have 
    $$
    - \one^T VW^T \one = \sum_{i=1}^d \inner{\one}{-a_i} \inner{b_i}{\one} = \sum_{i=1}^d \|\one\|^2 \|a_i\|\|b_i\| \cos(\phi_i) \cos(\psi_i). 
    $$
    Since the angle between $-a_i$ and $b_i$ is $\pi - \theta_i$, we have $\phi_i + \psi_i \geq \pi - \theta_i$. Since the cosine function is log-concave on $[0,\pi]$, we have 
    $$
    \cos(\phi_i) \cos(\psi_i) \leq \cos^2\left(\frac{\pi-\theta_i}{2}\right).
    $$
    By combining the equations above, we have
    \begin{align*}
    k \sum_{i=1}^d \|a_i\|\|b_i\| \cos^2\left(\frac{\pi-\theta_i}{2}\right) \geq - \one^T VW^T \one \geq - \Tr(VW^T) + k(k-1-L)\beta - kL.
    \end{align*}
    Since $\Tr(VW^T) \leq k$, we have
    \begin{equation}\label{eq:robust-all-sum}
        k \sum_{i=1}^d \|a_i\|\|b_i\| \cos^2\left(\frac{\pi-\theta_i}{2}\right) \geq -k+k(k-1-L)\beta - kL.
    \end{equation}

    By combining Equations~(\ref{eq:robust-trace}) and~(\ref{eq:robust-all-sum}), we have 
    $$
    \sum_{i=1}^d \|a_i\|\|b_i\| = \sum_{i=1}^d \|a_i\|\|b_i\| \left(2\cos^2\left(\frac{\pi-\theta_i}{2}\right) + \cos(\theta_i)\right) \geq k\alpha+(2k-2-2L)\beta -2 - 2L.
    $$
    By Cauchy-Schwarz inequality, we have 
    $$
    \left(\sum_{i=1}^d \|a_i\|\|b_i\| \right)^2 \leq \sum_{i=1}^d \|a_i\|^2 \cdot \sum_{i=1}^d \|b_i\|^2 = k^2,
    $$
    where the last equality is from $\sum_{i=1}^d \|a_i\|^2 = \sum_{i=1}^d \|b_i\|^2 = k$ since matrices $V$ and $W$ are both consists of $k$ unit row vectors. Therefore, we have 
    $$
    k \leq \frac{(2+2L)(\beta+1)}{\alpha + 2\beta-1}.
    $$

\end{proof}

\end{document}